\documentclass[letterpaper, 10 pt, conference]{ieeeconf} 
\IEEEoverridecommandlockouts 

\usepackage{cite}
\usepackage[pdftex]{graphicx}
    \graphicspath{{img/},{fig/}}
\usepackage{amsmath,amssymb}
\usepackage{array}
\usepackage[caption=false,font=footnotesize]{subfig}
\usepackage{fixltx2e}
\usepackage{siunitx}
\usepackage{threeparttable}

\usepackage{hyperref}
\usepackage[utf8]{inputenc}
\usepackage{gensymb}

\usepackage{xcolor}
\usepackage{soul}

\makeatletter
\newcommand*\titleheader[1]{\gdef\@titleheader{#1}}
\AtBeginDocument{%
  \let\st@red@title\@title
  \def\@title{%
    \bgroup\normalfont\small\raggedright\@titleheader\par\egroup
    \vskip0.5em\st@red@title}
}
\makeatother
\title{\LARGE \bf
Low-Level Force-Control of MR-Hydrostatic Actuators
}
\titleheader{2021 IEEE International Conference on Robotics and Automation (ICRA 2021). Preprint version. Accepted February 2021.$^{2}$} 

\begin{document}

\author{Jeff Denis$^{1}$, Jean-Sébastien Plante$^{1}$ and Alexandre Girard$^{1}$

\thanks{\scriptsize{This work was supported by the Fonds québécois de la recherche sur la nature et les technologies (FRQNT), the Ministère de l'Économie, de la Science et de l'Innovation (MESI), Exonetik Inc. and the Natural Sciences and Engineering Research Council of Canada (NSERC). \textit{(Corresponding author:
Jeff Denis.)}}}

\thanks{$^{1}$\scriptsize{All authors are with the Department of Mechanical Engineering, Université de Sherbrooke, Qc, Canada.}
        {\tt\scriptsize jeff.denis@usherbrooke.ca}}%
\thanks{$^{2}$\scriptsize{© 2021 IEEE. Personal use of this material is permitted. Permission from IEEE must be obtained for all other uses, in any current or future media, including reprinting/republishing this material for advertising or promotional purposes, creating new collective works, for resale or redistribution to servers or lists, or reuse of any copyrighted component of this work in other works. DOI:10.1109/LRA.2021.3063972}}
}


\maketitle

\begin{abstract}
Precise and high-fidelity force control is critical for new generations of robots that interact with humans and unknown environments. Mobile robots, such as wearable devices and legged robots, must also be lightweight to accomplish their function. Hydrostatic transmissions have been proposed as a promising strategy for meeting these two challenging requirements. In previous publications, it was shown that using magnetorheological (MR) actuators coupled with hydrostatic transmissions provides high power density and great open-loop human-robot interactions.

This paper compares control strategies for MR-hydrostatic actuator systems to increase its torque fidelity, defined as the bandwidth (measured vs desired torque reference) and transparency (minimizing the undesired forces reflected to the end effector when backdriving the robot). Four control approaches are developed and compared experimentally: (1) Open-loop control with friction compensation; (2) non-collocated pressure feedback; (3) collocated pressure feedback; (4) LQGI state feedback. A dither strategy is also implemented to smoothen ball screw friction. Results show that approaches (1), (2) and (3) can increase the performances but are facing compromises, while approach (4) can simultaneously improve all metrics. These results show the potential of using control schemes for improving the force control performance of robots using tethered architectures, addressing issues such as transmission dynamics and friction.

\end{abstract}

\section{Introduction}

Actuators and transmissions are key factors for the performance of a robotic system that interacts with humans and uncertain environments. This includes exoskeletons interacting with human limbs, robotic arms interacting with delicate objects and legged robots that interact with the ground. Transparency means that the actuator is backdrivable or delivers the expected force and impedance without parasitic dynamics \cite{carignan_closed-loop_2000} \cite{buerger_novel_2010}.

A perfect actuator and transmission system would always apply the desired force on the environment. However, internal dynamics and friction create parasite forces that lead to deviations from the desired force level. Actuators with low level of parasite forces are highly backdrivable. This quality is referred as transparency \cite{carignan_closed-loop_2000} \cite{buerger_novel_2010}. Also, real actuators cannot respond to a desired torque at high frequencies. This limitation, a consequence of internal dynamics, is described by the control bandwidth, a maximum frequency level above which the actuation system is unable to produce the desired force. A perfect actuator and transmission system would have zero output intrinsic impedance (perfect transparency) and infinitely high control bandwidth \cite{zinn_new_2004}. The overall capacity to apply exactly the desired force level on the environment despite the conditions will be called here the force fidelity.

Hydrostatic transmissions have already proven to be a promising technology for lightness and transparency. Indeed, rolling diaphragms and water transmissions provide low friction and low inertia for applications with human-human interactions and manual tele-operation \cite{whitney_hybrid_2016} \cite{burkhard_rolling-diaphragm_2017}. In contrast with cable transmissions, hydrostatic transmissions are easy to route thus permitting more flexibility for robotics system design.

Hydrostatic transmissions can be combined with transparent actuators for high torque fidelity. For robotic devices, direct drive motors have been used with hydrostatic transmissions for human-robot interactions such as for a collaborative gripper \cite{schwarm_floating-piston_2019} and for an elbow exoskeleton \cite{bolignari_design_2020}. As a lightweight and compact alternative to direct drive motors, magnetorheological (MR) actuators were used in a supernumerary hydrostatic robotic leg \cite{khazoom_supernumerary_2020} and an ankle hydrostatic exoskeleton \cite{khazoom_design_2019}. MR actuators have good torque-to-weight ratios while having a low intrinsic impedance, no cogging and no backlash.
MR actuators consist of a highly geared motor paired with an MR clutch so that the motor inertia is decoupled from the output. MR actuators have high torque bandwidth and torque-to-inertia ratio \cite{viau_tendon-driven_2017}.

\begin{figure}[t!]
\centering
\includegraphics[width=0.4\textwidth]{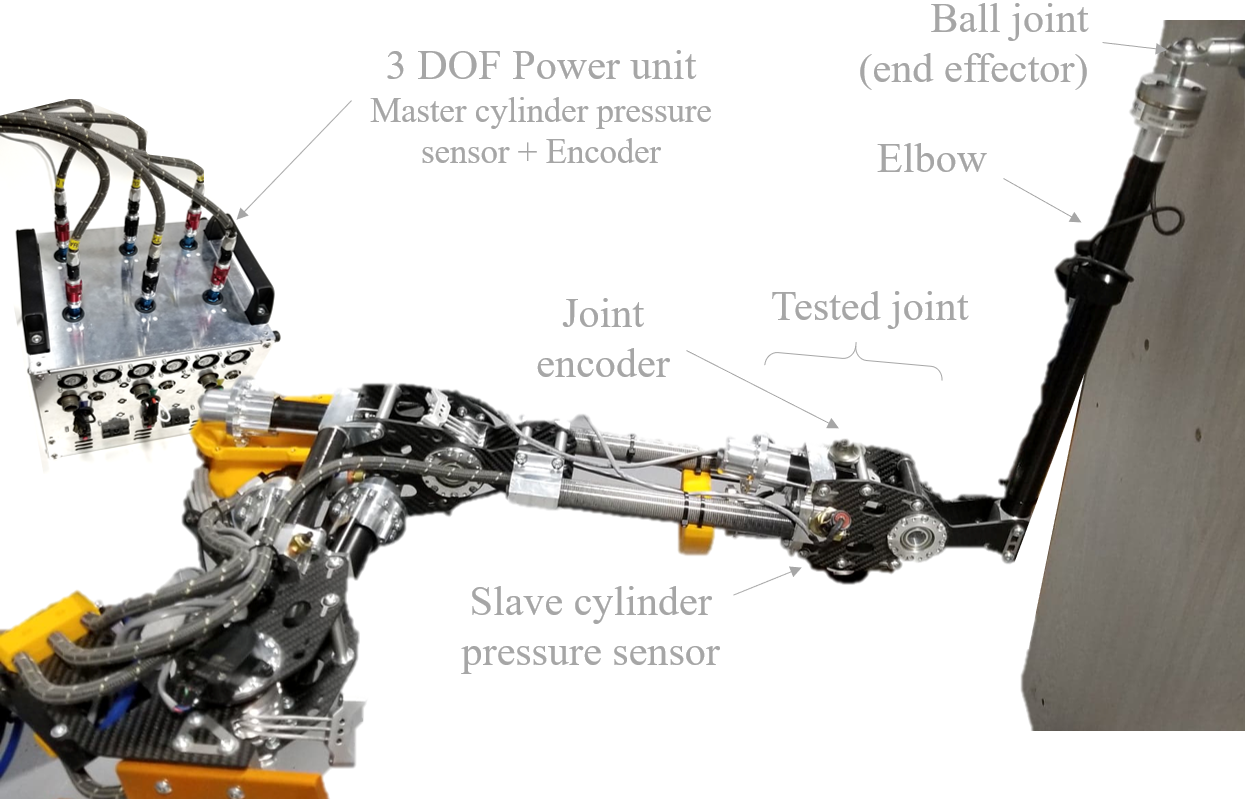}
\caption{The MR-hydrostatic supernumerary robotic arm prototype as set in this paper for experimental torque fidelity evaluation.}
\label{fig_TestBench}
\end{figure}

In \cite{veronneau_multifunctional_2020}, a three degrees of freedom (DOF) supernumerary robotic arm with a mass of 4.2~kg and using MR actuators was shown to have a 5~kg payload capacity (Fig.~\ref{fig_TestBench}). The 18~kg delocalized power unit features six MR clutches and hydrostatic transmission lines. Low friction rolling diaphragms seals the pistons. High-pitch ball screws are used to convert MR output torque to linear force on piston. The prototype showed good torque density, backdrivability and a torque bandwidth up to 18~Hz limited by the hydrostatic transmission natural frequency due to the presence of dissolved air, sealing flexibility, hose length and internal diameter along with the inertia of the actuator \cite{veronneau_high-bandwidth_2018}.  

This paper aims to compare different low-level control approaches to increase torque fidelity of the proposed MR-hydrostatic actuator, compensating for the ball screw nonlinear friction and the effect of transmission’s dynamics. The following results show the potential to improve force control performances of any tethered actuator architectures which does not necessarily include MR clutches.

Section~II reviews the related works regarding friction reduction and high bandwidth control of compliant transmissions. Section~III briefly describes the mechanical design of the test prototype. Section~IV proposes a dynamic model of MR-hydrostatic actuators. Section~V introduces four controllers for improving torque fidelity: a dither-based approach for smoothing friction, a simple friction compensation technique, typical pressure feedback strategies and finally a state feedback approach. Then, section~VI compares and discusses the experimental results obtained for a single hydrostatic line of the robotic arm in blocked output.

\section{Control for friction and compliance}

\subsection{Friction reduction}
Ball screws are a torque dense device to convert actuator torque to linear force. However, as lead screws, ball screws exhibit Coulomb friction and stiction \cite{juvinall_fundamentals_2017}.

Many model-based approaches as well as disturbance observers exist to estimate and compensate non-linear friction. In \cite{wang_series_2020}, a disturbance observer is used to effectively reduce motor friction in a hydrostatic actuator based on pressure measurements and differentiating position measurements to obtain acceleration. However, high frequency disturbance rejection is limited by the resolution of the position sensor or by EMI noise. Sensor fusion techniques using an accelerometer and position sensor may be used to increase the range of disturbance rejection.

In \cite{seok_actuator_2012}, the Coulomb friction in a leg actuator is compensated based on the displacement of the leg during the walking pattern. This strategy is simple and efficient for fast disturbances such as impacts. A similar approach is used for a hip assistance ball screw driven device.  \cite{olivier_ball-screw_2014} However, at low speeds, stiction can create additional parasitic forces (stick-slip effect) and can also limit speed tracking performances.

To break stiction, a simple method is to generate a high frequency vibration at the friction point to ensure slippage condition at all times. In \cite{chen_ballscrew_2000}, adding a piezoelectric actuator on the lead screw nut reduces the friction torque under different load conditions from 20\% to 45\% and smoothens the discontinuity of friction during reversals. An analog strategy is proposed in this paper but directly using the MR clutches.

\subsection{Control beyond transmission resonance}
\label{section_controlBeyondResonance}
In most robotic devices, closing the loop with a force sensor may be a solution to increase performance. However, hydrostatic transmissions are compliant. With non-collocated force-control (sensor far from the actuator), instability will occur when increasing controller gains. Collocated force control (sensor near actuator) can allow increasing feedback gains but it can still excite flexible modes. \cite{eppinger_three_1992} \cite{veronneau_lightweight_2019}


Applying a notch filter to the controller output is a simple but limited approach in open-loop controls to damp a resonance without minimal loss of control bandwidth. However, accuracy and disturbance rejection is limited in open-loop control. An alternative approach is to actively damp the structural modes by pole placement. For example, pole placement control can be used to increase high frequency disturbance rejection and damp the flexible modes for ball screw drives in CNC machines. These techniques require a way to measure or observe all states related to flexible modes. \cite{gordon_accurate_2013} \cite{dumanli_optimal_2018}

By extension, pole placement techniques could be used with hydrostatic transmissions to damp the high frequency poles and increase the reference tracking performances. This is done in Section~V with the state feedback controller.

\section{Mechanical design}

This section briefly presents the robotic actuator used in this work. Some details regarding energy consumption of the actuator for wearable robotics are also presented to complement the information of the previous papers \cite{veronneau_lightweight_2019} \cite{veronneau_multifunctional_2020}.

\begin{figure*}[t!]
\centering
\includegraphics[width=0.99\textwidth]{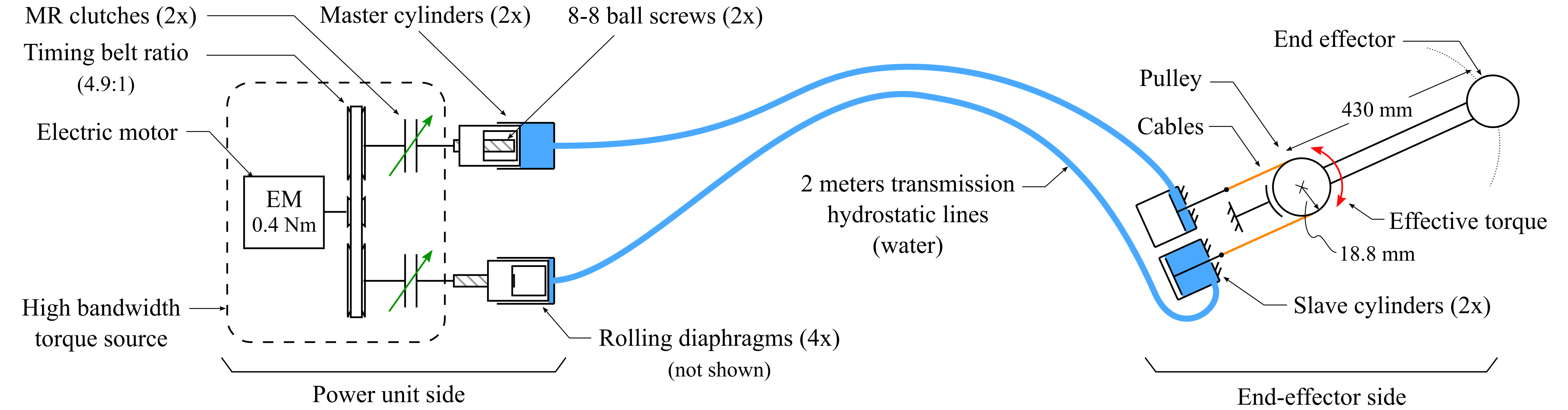}
\caption{Schematic drawing (1 DOF) of the actuator used.}
\label{fig_Kinematics}
\end{figure*}

The system consists of a high bandwidth and backdrivable positive pressure source whose mechanical power is transmitted to robotic joints by means of low-friction hydrostatic water lines. Two lines are used for each robotic joint in an antagonistic manner. Since the lines are independent and positive pressures are used, the design does not exhibit any backlash. The mechanical system for one DOF is illustrated in Fig.~\ref{fig_Kinematics}.

For one DOF, the power unit comprises an inexpensive geared high-speed BLDC motor (KDE600XF-1100-G3, KDEDirect) to power two custom 2~\si{N.m} MR~clutches~(0.73~\si{kg} each). Here, the motor is controlled with a simple open loop voltage control method. Ball screws (GTR0808EC5T-220, ISSOKU, 8~\si{mm} diameter and lead) converts the clutch torque to pressure by means of master pistons. The ratio $R$ between joint torque and clutch torque is 14.7 so the maximum torque generated at the joint is 29~\si{N.m} (at a pressure of 2310~\si{kPa} or 335~\si{psi}). The torque density of the power unit is 10~\si{N.m/kg} (excluding transmission), which is high compared to direct-drive motors providing similar force fidelity. The output inertia of each clutch and screw rod is low (0.15~\si{kg.cm^2} and 0.02~\si{kg.cm^2} respectively), resulting in a low reflected mass when backdriving the end effector. The pistons are sealed with rolling diaphragms to minimize friction. The power is transmitted to the robotic joint by means of hydrostatic water lines. On the end-effector side, slave pistons transmit the force to the robotic joint by means of steel cables.

As shown in table~\ref{table_PowerConsumption}, in most cases, MR clutches slightly increase power consumption if compared to the same design without MR clutches (i.e. one motor driving two ball screws directly). However, for the specific case of high torque applied and blocked output, power consumption is higher with clutches because the motor must maintain a constant speed at high torque. For the same condition, the power consumption of direct drive motors mostly consists of Joule losses. Power calculations for MR actuators include Joule dissipation in motor and clutch coils and required motor mechanical power. A 10~\si{rad/s} controlled MR slip speed was used. The maximum power case is 11.5~\si{N.m} at 6.3~\si{rad/s} at the joint as achieved in \cite{veronneau_multifunctional_2020}.

\begin{table}[h!]
\vspace{10pt}
	\centering
		\vspace{-10pt}
	\caption{Power consumption (W) analysis for 3 operation points}
		\begin{tabular}{ l c c }
		    
			Operation point & Motor only & Motor + clutches  \\ \hline
			No torque and no speed & 0 & 0.3\\ \hline 
			Max torque with blocked output & 18 & 65 \\ \hline
			Max power achieved & 76 & 90 \\ \hline
		\end{tabular}
	\label{table_PowerConsumption}
		\vspace{-5pt}
\end{table}

\section{Modeling and characterization}
\label{section_modeling}

\subsection{MR clutches}
\label{section_MRclutches}
\begin{figure}%
    \centering
    \subfloat[\centering]{{\includegraphics[width=0.2\textwidth]{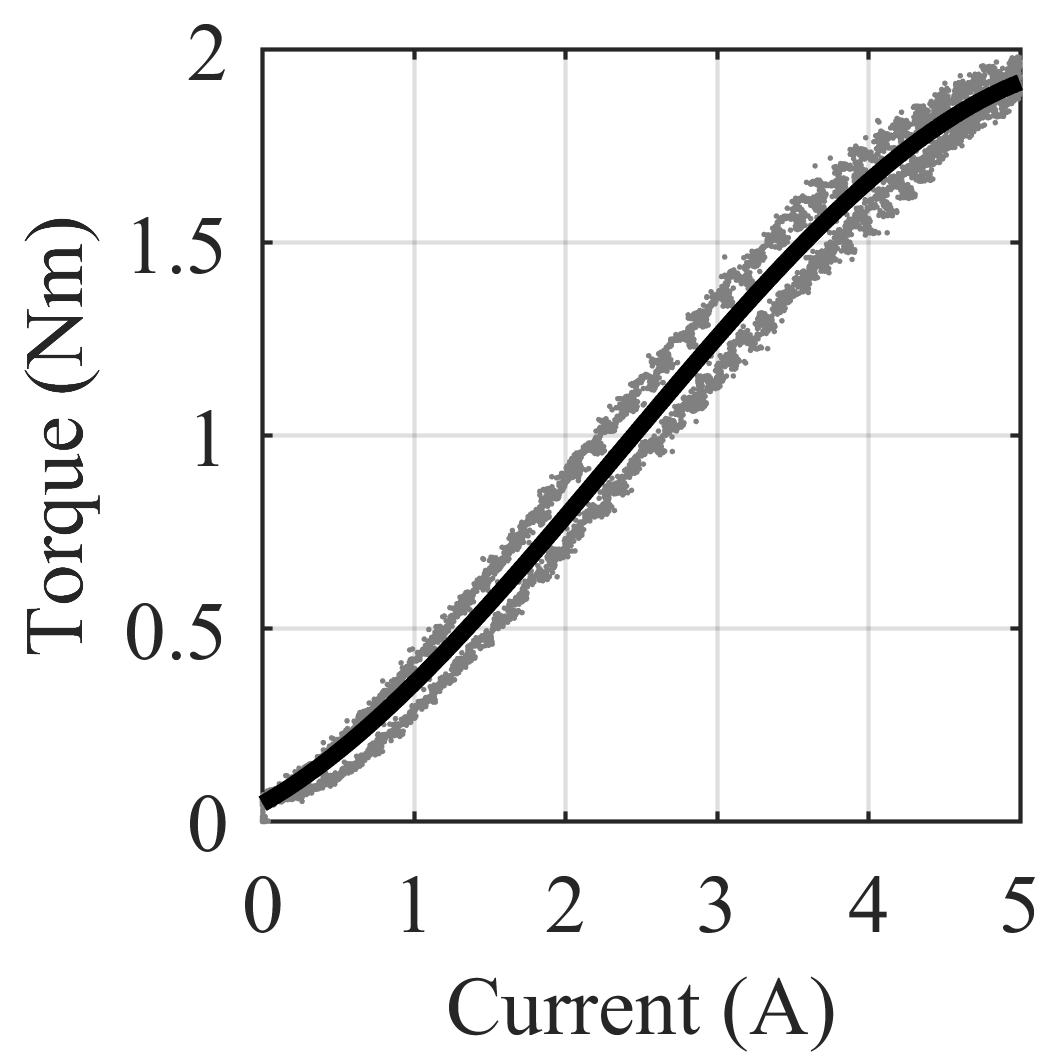}}}%
    \qquad
    \subfloat[\centering]{{\includegraphics[width=0.2\textwidth]{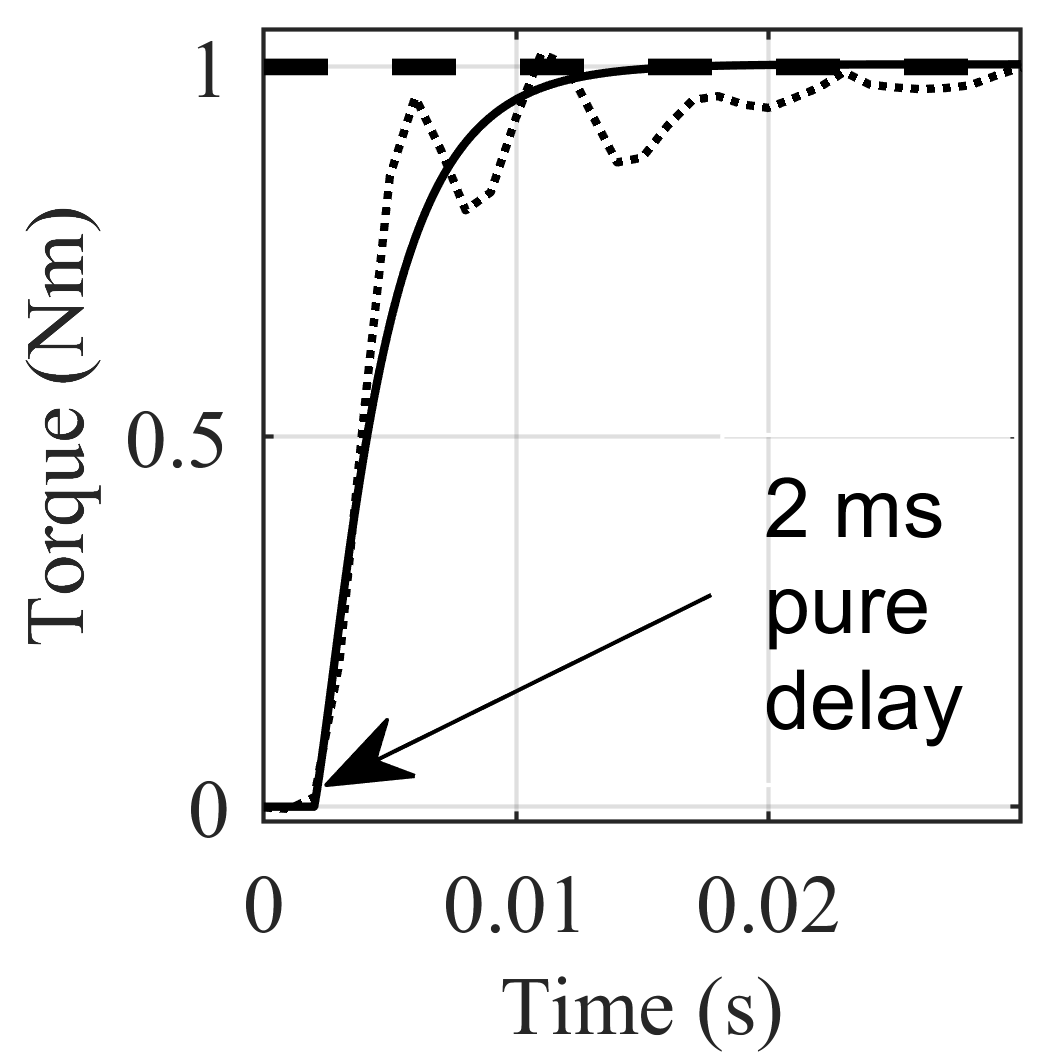}}}%
    \caption{MR clutches characterization curves: a) output torque as a function of the input current for five successive up-down current ramps (grey dot) and polynomial fit (solid black); b) normalized torque step response (grey dot) to a 2.5 A command (dash) with about 100~\si{rad/s} slip speed and a first-order model with pure delay model (solid black).}%
    \label{fig_MRcharacterizations}%
\end{figure}

The static MR torque curve is presented in Fig.~\ref{fig_MRcharacterizations}a. 
The data describes five successive 20 seconds up-down current ramps. The slip speed is about 100~\si{rad/s}. A torque cell (TRS605-20Nm, Futek)  measures the output torque. A nonlinear 10\% hysteresis based on maximum torque is obtained and is due to the ferromagnetic core of MR clutches \cite{yadmellat_adaptive_2016}. This behavior is not modeled for simplicity. Equation~(\ref{eq_MR}) is a third order polynomial curve fit that can be used to convert a desired torque to a current command~:
\begin{equation}
    {{T}_{\text{MR}}}=-0.015{{I}_{\text{MR}}}^{3}+0.104{{I}_{\text{MR}}}^{2}+0.225{{I}_{\text{MR}}}+0.044
    \label{eq_MR}
\end{equation}

The MR output torque also depends on the slip speed and direction. The MR viscosity found is 0.35~\si{mN.m.s/rad} and the static friction is 15~\si{mN.m}. Since the MR clutches work in an antagonistic manner, static friction torque of each clutch cancels at output. Also, if the slip speed (i.e. the speed difference between clutch input and output) of each clutch is equal, the viscous friction also cancels at the output.

To assess the dynamic behavior of the MR clutch, a 2.5~\si{A} step command is sent to a clutch. Fig.~\ref{fig_MRcharacterizations} presents the torque response obtained. A first-order model describes the torque response well and is given by equation~(\ref{eq_ClutchDynamics}).

\begin{equation}
    {{H}_{\text{MR}}}(s)\approx {{e}^{-{{\tau }_{\text{MR}}}s}}\frac{{{\omega }_{\text{c}\text{,MR}}}}{s+{{\omega }_{\text{c}\text{,MR}}}}
    \label{eq_ClutchDynamics}
\end{equation}

where the pure delay due to the drive and MR fluid response is ${{\tau }_{\text{MR}}}=0.002$~\si{\second} and the cutoff frequency is ${{\omega }_{\text{c}\text{,MR}}}=2\pi64$~\si{rad/s}.

\subsection{Transmission model}
\label{subsection_Transmission}
\begin{figure}[h!]
\centering
\includegraphics[width=0.45\textwidth]{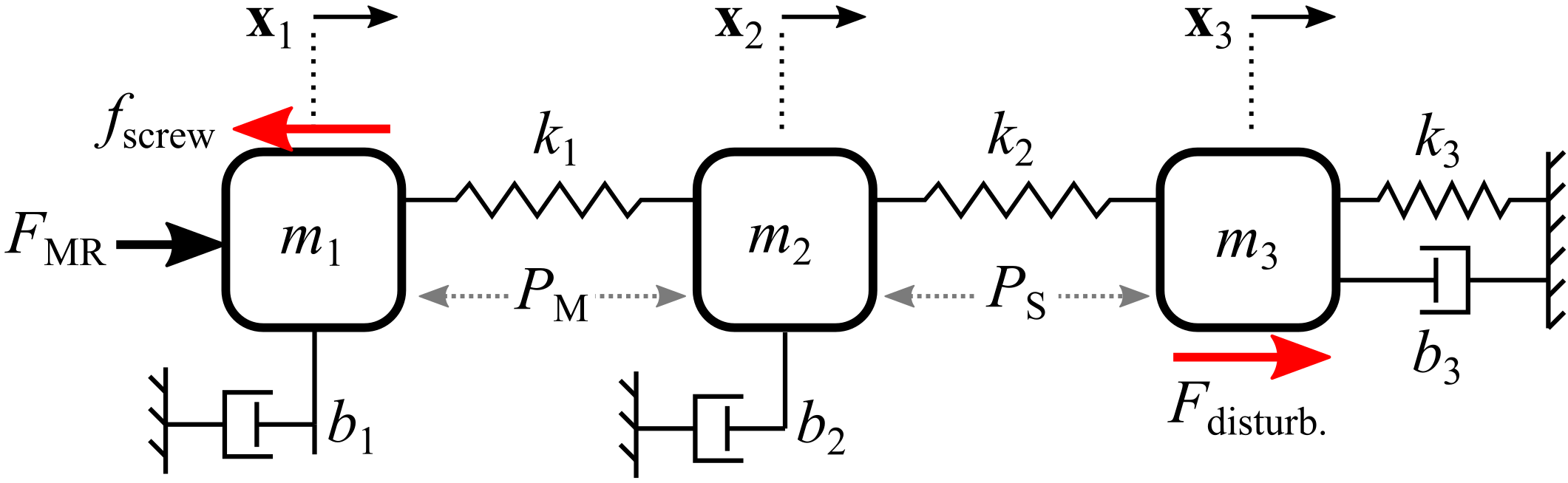}
\caption{Modelisation of the hydrostatic  transmission.}
\label{fig_Model}
\end{figure}

The same sixth-order model as developed in \cite{veronneau_lightweight_2019} is used to represent the linear dynamics of the hydrostatic transmission in series with the MR clutch and the robot mass (Fig.~\ref{fig_Model}). The compliance of the transmission is mainly due to the rolling diaphragms and the dissolved air in water. This compliance varies with pressure and was calculated for a 900~\si{kPa} (130~\si{psi}) mean pressure. All parameters of the model are presented in table~\ref{table_parameters} and given as reflected in the slave piston referential which explains the large magnitudes.

\begin{table}[h!]
\renewcommand{\arraystretch}{1.1}
\caption{MR-hydrostatic actuation model parameters}
\label{table_parameters}
\centering
\begin{tabular}{ c l c c }
 Parameter & Description (units) & Value \\
\hline
$m_{1}$ & Clutch + ball screw + piston mass (\si{kg}) & 11 \\
$k_{1}$ & Power unit transmission stiffness (\si{N/m}) & \num{6.2e5} \\
$b_{1}$ & MR clutch + ball screw damping (\si{N.s/m}) & 650 \\
$m_{2}$ & Hydraulic fluid mass (\si{kg}) & 7  \\
$k_{2}$ & Robot side transmission stiffness (\si{N/m}) & \num{5.3e5} \\
$b_{2}$ & Hydraulic viscous damping (\si{N.s/m}) & 204 \\
$m_{3}$ & Robot structure + payload mass (\si{kg}) & 976 \\
$k_{3}$ & Robot structure + base stiffness (\si{N/m}) & \num{2.2e5} \\
$b_{3}$ & Robot + base viscous damping (\si{N.s/m}) & 10000  \\
\hline
\end{tabular}
\end{table}

Parameters $m_3$, $k_3$ and $b_3$ represents the overall impedance of the robotic structure, base (human hips or test bench) and wall when the robotic arm interacts with the environment at the end effector. Indeed, the robot is not perfectly rigid and $k_3$ and $m_3$ were estimated by comparing experimental joint displacement and acceleration with the measured joint torque. The model was also validated by sending a chirp command (0–200~Hz, $2\pm0.5$~A) to a clutch while the end effector is blocked. The torque frequency response function (FRF) of the prototype was found. Some tuning for the viscous damping parameters and $m_2$ was necessary to get a better fit with experimental data. The model and experimental FRF are given in Fig.~\ref{fig_BodeResponsesControllers}.

\subsection{Nonlinear friction}

The nonlinear friction due to the ball screw introduces disturbance between the clutch output and the master piston. Hence, the slave pressure $P_{\text{S}}$ changes with the direction of movement when the MR clutch delivers a constant torque. This must be compensated to provide good torque fidelity. A simple model for the ball screw is a Coulomb friction model whose amplitude is proportional to the load on the nut:
\begin{equation}
   {P}_{\text{f}}=\mu {{P}_{\text{nut}\text{,load}}}\text{sign}\left( {{{\dot{x}}}_{1}} \right)
    \label{eq_ScrewFriction1}
\end{equation}

where ${P}_{\text{f}}$ is the pressure deviation due to friction and where $P_{\text{M}}$ is the pressure in the master cylinder. Also, to avoid discontinuity around zero speed, a hyperbolic tangent function is used to replace the sign function, which gives:

\begin{equation}
    {{P}_{\text{f}}}=\mu {{P}_{\text{M}}}\tanh \left( n{{{\dot{x}}}_{1}} \right)
    \label{eq_ScrewFriction2}
\end{equation}

The parameter $n$ in (\ref{eq_ScrewFriction2}) is used to adjust the steepness of the transition. To find the friction parameter $\mu$, the MR-hydrostatic actuator is backdriven with a 1~Hz position sine wave. The maximum piston speed is 5~mm/s. The torque delivered by the clutch is increased slowly. Hence, the mean hydrostatic pressure increases as well as the deviation due to the ball screw friction. From the experimental data, the pressure difference from the nominal pressure is found and represents the pressure ${P}_{\text{f}}$ resulting from the friction. The linear fit reveals $\mu=0.14$ with a coefficient of determination of 0.981.

\section{Controller design}
This section develops various control strategies to improve torque fidelity of the MR-hydrostatic actuator system. Four classic control approaches are considered in increasing level of complexity and discussed below:

\begin{enumerate}
    \item Open-loop control with friction compensation (Fig.~\ref{fig_OpenLoop_CTRL})
    \item Non-collocated pressure feedback (Fig.~\ref{fig_NonCollocated_CTRL})
    \item Collocated pressure feedback (Fig.~\ref{fig_NonCollocated_CTRL})
    \item LQGI state feedback (Fig.~\ref{fig_LQGI_CTRL})
\end{enumerate}

For all these controllers, a dither signal is implemented to smoothen nonlinear friction in the ball screw.

\subsection{Open-loop controller with friction compensation}
\label{subsection_openloopCTRL}
In open-loop, the MR clutches are controlled without any force/pressure feedback and without considering transmission dynamics. Only ball screw friction is compensated. Fig.~\ref{fig_OpenLoop_CTRL} shows the controller:

\begin{figure}[h!]
\centering
\includegraphics[width=0.49\textwidth]{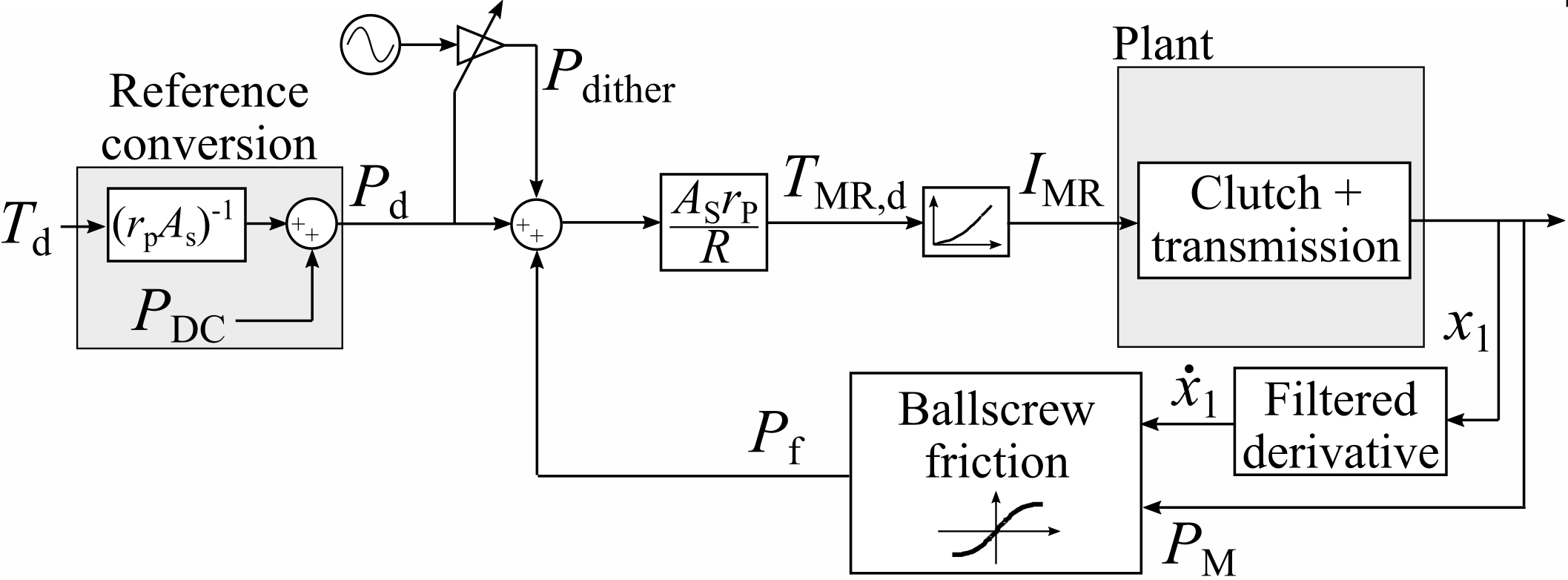}
\caption{Open-loop controller with simple ball screw friction compensation diagram and the implementation of the dither command.}
\label{fig_OpenLoop_CTRL}
\end{figure}

First, the desired torque at joint $T_\text{d}$ is converted to a desired pressure $P_\text{d}$ at the slave cylinder using the cylinder area $A_\text{s}$ and the joint pulley radius $r_\text{p}$. A DC pressure $P_\text{DC}=$205~kPa (30~psi) is added to the reference so that a tension in the cables is maintained. In an antagonistic joint, the DC pressure commands produce a zero effective torque at the joint. Then, the desired pressure is converted to a desired MR linear force. The required MR coil current is found using a lookup table from equation~\ref{eq_MR}.

The ball screw friction is compensated by:
\begin{enumerate}
    \item breaking the discontinuous stiction effect with a dither command
    \item compensating the friction now that the friction behavior is continuous over the speed range
\end{enumerate}
MR clutches directly generate a high frequency sine wave pressure command $P_{\text{dither}}$. A dither frequency of 150~\si{Hz} is chosen as a compromise where MR clutches can still deliver torque amplitude while hydrostatic transmission mostly damps pressure waves. The dither amplitude is increased linearly with the desired pressure. For the friction compensation, equation~(\ref{eq_ScrewFriction2}) is implemented to estimate the friction and feed it to the command. The ball screw speed ${{\dot{x}}_{1}}$ is differentiated from the measured ball nut position with a 150~Hz low-pass filter.

\subsection{Non-collocated and collocated feedback controllers}
\label{subsection_PressureFeedbackCTRL}
\begin{figure}[h!]
\centering
\includegraphics[width=0.49\textwidth]{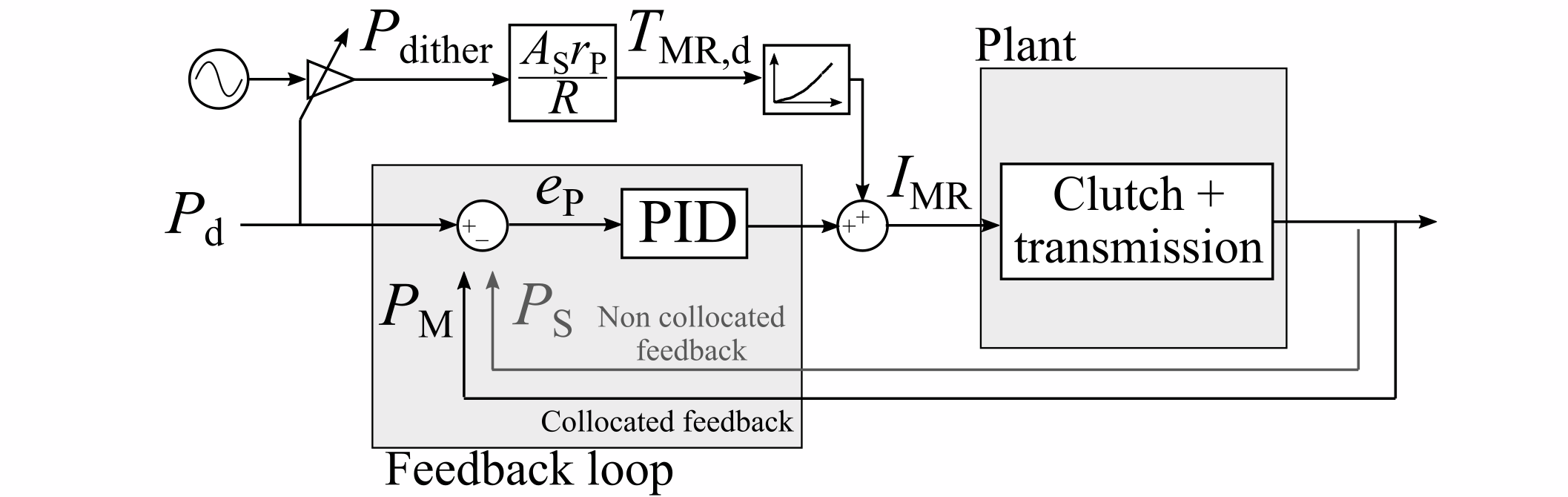}
\caption{Closed-loop controllers with feedback either on master cylinder pressure (collocated control) or on slave cylinder pressure (non-collocated control). The reference conversion from a desired torque $T_\text{d}$ to a desired pressure $P_\text{d}$ is omitted here for simplicity.}
\label{fig_NonCollocated_CTRL}
\end{figure}

Simple non-collocated and collocated controllers are developed for comparison purposes (Fig.~\ref{fig_NonCollocated_CTRL}). A feedback loop based on the slave and master pressure measurements, respectively, is used with the dither signal. High controller gains typically increase closed-loop bandwidth and disturbance rejection. However, as discussed in section~\ref{section_controlBeyondResonance}, transmission dynamics limits the aggressiveness of these controllers. Indeed, based on the model presented in section~\ref{subsection_Transmission}, the amplitude of the resonant modes of the loop transfer function for both controller limits the achievable closed-loop bandwidth due to gain margin issues. Based on loop shaping, most part of the appropriate PID controllers is just integral action with limited gain for stability. This is what is used for experimental validations.

Still, shaping the loop transfer functions at the resonant modes is possible with higher order controllers. The maximum closed-loop bandwidth would then be closer to the open-loop bandwidth. In the next subsection, state feedback is developed to push performance further.


\subsection{State feedback}
\begin{figure}[h!]
\centering
\includegraphics[width=0.49\textwidth]{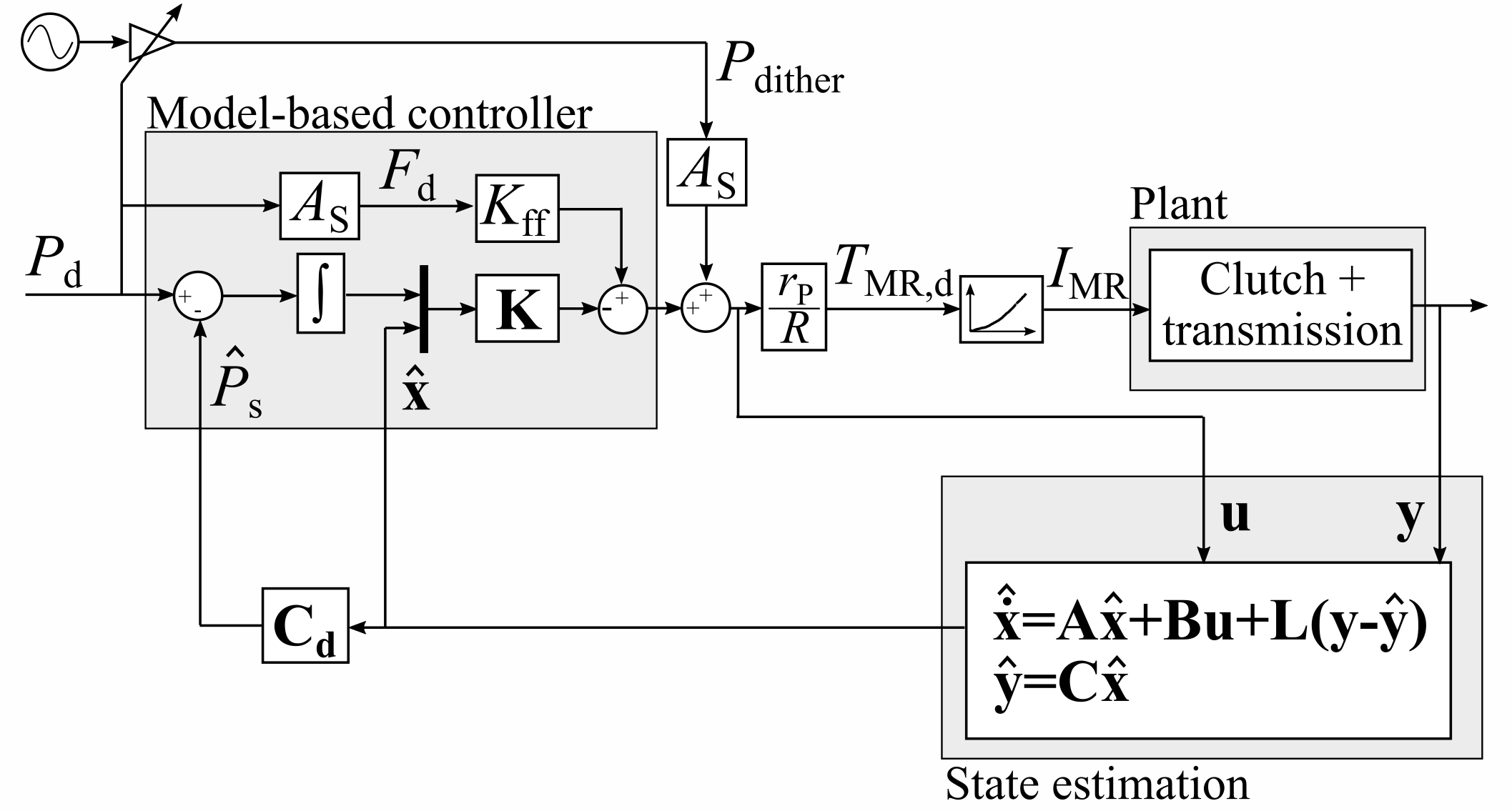}
\caption{LQGI controller diagram for the MR-hydrostatic actuator. The regulator $\mathbf{K}$ stabilizes the estimated states and the integral error of the estimated slave pressure. A feedforward gain $K_{\text{ff}}$ generates the appropriate command to achieve the desired slave pressure based on the linear model. The Kalman filter uses the estimated force, the measurements and the model to output the estimated states. The reference conversion from a desired torque $T_\text{d}$ to a desired pressure $P_\text{d}$ is omitted here for simplicity.}
\label{fig_LQGI_CTRL}
\end{figure}

State feedback manages the hydraulic resonance and pushes the bandwidth of the system up to the physical limits of the system. A Linear-Quadratic-Gaussian-Integral controller (LQGI) achieves this goal. The integral action deals with ball screw friction. The controller diagram is presented in Fig.~\ref{fig_LQGI_CTRL}.

The following state-space model represents the system described in sections III and IV:

\begin{equation}
    \mathbf{\dot{x}}=\mathbf{Ax}+\mathbf{Bu}\
\end{equation}

where\\

\noindent
\begin{equation}
    \resizebox{0.91\hsize}{!}{$
   \textbf{A}={{\left[ \begin{matrix}
   0  \\
   \frac{1}{{{m}_{1}}}  \\
   0  \\
   \frac{1}{{{m}_{2}}}  \\
   0  \\
   \frac{1}{{{m}_{3}}}  \\
   0  \\
\end{matrix} \right]}^{\text{T}}}\left[ \begin{matrix}
   0 & 1 & 0 & 0 & 0 & 0 & 0  \\
   -{{k}_{1}} & -{{b}_{1}} & {{k}_{1}} & 0 & 0 & 0 & 1  \\
   0 & 0 & 0 & 1 & 0 & 0 & 0  \\
   {{k}_{1}} & 0 & -{{k}_{12}} & -{{b}_{2}} & {{k}_{2}} & 0 & 0  \\
   0 & 0 & 0 & 0 & 0 & 1 & 0  \\
   0 & 0 & {{k}_{2}} & 0 & -{{k}_{23}} & -{{b}_{3}} & 0  \\
   0 & 0 & 0 & 0 & 0 & 0 & -{{\omega }_{\text{c},\text{MR}}}  \\
\end{matrix} \right] $}
\end{equation}

\begin{equation}
\textbf{B} = {{\left[ \begin{matrix}
   0 & 0 & 0 & 0 & 0 & 0 & {{\omega }_{\text{c}\text{,MR}}}  \\
\end{matrix} \right]}^{\text{T}}}
\end{equation}

where $k_{12}=k_1+k_2$ and $k_{23}=k_2+k_3$. The state vector $\mathbf{x}$ and the input vector $\mathbf{u}$ are:

\begin{equation}
\mathbf{x}={{\left[ \begin{matrix}
   {{x}_{1}} & {{{\dot{x}}}_{1}} & {{x}_{2}} & {{{\dot{x}}}_{2}} & {{x}_{3}} & {{{\dot{x}}}_{3}} & {{F}_{\text{MR}}}  \\
\end{matrix} \right]}^{\text{T}}}
\end{equation}
\begin{equation}
\mathbf{u}=\left[ {{F}_{\text{MRs}}} \right]
\end{equation}
where $x_i$ are the displacements of the model’s masses, ${{F}_{\text{MR}}}$ is the MR force and ${{F}_{\text{MRs}}}$ is the steady-state MR force expected for a given current based on the characterization. The state-space model includes the first-order dynamics of the clutch based on equation~(\ref{eq_ClutchDynamics}). However, the linear model omit the ball screw nonlinear friction and the MR pure delays. The desired output to track $y_{\text{d}}$ is the slave pressure and is given by equation~(\ref{eq_y_Cx}):
$$y_{\text{d}}=\mathbf{C_dx}$$
\begin{equation}
   {y_{\text{d}}=\left[ \begin{matrix}
   0 & 0 & {{{k}_{2}}}/{{{A}_{\text{s}}}}\; & 0 & {-{{k}_{2}}}/{{{A}_{\text{s}}}}\; & 0 & 0  \\
\end{matrix} \right]\mathbf{x}}
\label{eq_y_Cx}
\end{equation}

The state-feedback law is:
\begin{equation}
    {\mathbf{u}}=-\mathbf{K\mathbf{z}}+K_{\text{ff}}P_{\text{d}}
\end{equation}
where $\mathbf{K}$ is the regulator matrix, $\mathbf{z} = \begin{matrix} [ x_{i} & \mathbf{x}^{\text{T}} ]^{\text{T}} \end{matrix}$ is the state vector augmented with the integral error of the slave pressure, $K_{\text{ff}}$ is the feedforward gain and $P_{\text{d}}$ is the desired pressure. $\mathbf{K}$ minimizes a standard cost function given by equations~(\ref{eq_Jcost}) and (\ref{eq_costFunction}) which penalizes the slave pressure, the integral state and the input force, respectively.

\begin{equation} \label{eq_Jcost}
    J=\int\limits_{0}^{\infty }{\left( {{\mathbf{z}}^{\text{T}}}\mathbf{Qz}+{{\mathbf{u}}^{\text{T}}}\mathbf{Ru} \right)dt}
\end{equation}

\begin{equation}
    J=\int\limits_{0}^{\infty }{\left[ {{\left( {{y}_{\text{d}}} \right)}^{2}}+{\rho_{\text{i}}}{{\left( {{x}_{\text{i}}} \right)}^{2}}+{\rho}{{\left( {{F}_{\text{MRs}}} \right)}^{2}} \right]dt}
    \label{eq_costFunction}
\end{equation}

The gains $\rho$ and $\rho_{\text{i}}$ control the controller’s aggressiveness. They were tuned in simulations and experimentally to \num{1e-4} and 1000, respectively. The feedback gain \textbf{K} is obtained by solving the steady-state solution of the algebraic Riccati equation using the Matlab function \textit{lqi}. The following equation from \cite{astrom_feedback_2008} gives the feedforward gain $K_{\text{ff}}$:

\begin{equation}
    {{K}_{\text{ff}}}={-1}/{\left( \mathbf{C_{d}}{{\left( \mathbf{A}-\mathbf{BK} \right)}^{-1}}\mathbf{B} \right)}\
\end{equation}

Most of the states are not directly measured but a Kalman filter can estimate them:

\begin{equation} \label{eq_kalmanX}
   \mathbf{\hat{\dot{x}}}=\mathbf{A\hat{x}}+\mathbf{Bu}+\ \mathbf{L}\left( \mathbf{y}-\mathbf{\hat{y}} \right)
\end{equation}
\begin{equation} \label{eq_kalmanY}
     {{\mathbf{\hat{y}}}={{\mathbf{C}}}\mathbf{\hat{x}}}
\end{equation}
where $\mathbf{L}$ is the filter feedback gain, ${\mathbf{y}}$ the measurements and ${\mathbf{\hat{y}}}$ the estimated outputs. Then:

\[\mathbf{C}=\left[ \begin{matrix}
   1 & 0 & 0 & 0 & 0 & 0 & 0  \\
   0 & 1 & 0 & 0 & 0 & 0 & 0  \\
   0 & 0 & 0 & 0 & 1 & 0 & 0  \\
   {{{k}_{1}}}/{{{A}_{\text{M}}}}\; & 0 & {-{{k}_{1}}}/{{{A}_{\text{M}}}}\; & 0 & 0 & 0 & 0  \\
\end{matrix} \right]\]
\begin{equation}
 \mathbf{y}={{\left[ \begin{matrix}
   {{x}_{1}} & {{{\dot{x}}}_{1}} & {{x}_{\text{3}}} & {{P}_{\text{M}}}  \\
\end{matrix} \right]}^{\text{T}}}
\end{equation}

where ${x}_{1}$ and $\dot{x}_{1}$ (differentiated) are the master cylinder position and speed, ${x}_{3}$ is the joint position, ${P}_{\text{M}}$ is the master cylinder pressure and ${A}_{\text{M}}$ is the master cylinder area. Note that the robotic arm does not require a slave pressure sensor or any force sensor. It is later measured experimentally for validation purposes only. As for the regulator, $\mathbf{L}$ is obtained by introducing a cost function which matrices $\mathbf{R}_{\text{L}}$ and $\mathbf{Q}_{\text{L}}$ represent the covariance matrices of sensors’ noise and model’s states respectively. They are:

\begin{equation}
    {{\mathbf{R}}_{\text{L}}}=E\left( v{{v}^{\text{T}}} \right),\ {{\mathbf{Q}}_{\text{L}}}={{\rho }_{\text{L}}}{{\mathbf{D}}_{n\times n}}
\end{equation}
where $\mathbf{v}$ is a column vector of the sensors' noise, ${{\mathbf{D}}_{n\times n}}$ is a diagonal matrix and $n$ is the number of states. It is assumed that the measurements are white Gaussian non-correlated signals. Based on measurements at rest, ${{\mathbf{R}}_{\text{L}}}$ is a diagonal matrix whose diagonal elements are respectively \num{3.6e-9}, \num{1e-6}, \num{2.5e-11} and \num{5.6e5}. For ${{\mathbf{Q}}_{\text{L}}}$, the gain ${\rho }_{\text{L}}$ is used to tune how much the model represents the real system. A value of \num{3e-5} is found appropriate experimentally. For ${{\mathbf{D}}_{n\times n}}$, an identity matrix was first used. Then, for better state estimation when the arm is backdriven, the element ${{\mathbf{D}}_{2\times 2}}$ was changed to \num{1e5} and the element ${{\mathbf{D}}_{6\times 6}}$ was changed to \num{1e6}. The filter feedback gain $\mathbf{L}$ is obtained from the steady-state solution of the algebraic Riccati equation using $\mathbf{R}_{\text{L}}$, $\mathbf{Q}_{\text{L}}$ and the \textit{lqr} Matlab function.

\section{Experimental results and discussion}
To increase torque fidelity of hydrostatic transmissions, a fast and accurate torque response is desired in any situation. Hence, an ideal controller would increase torque bandwidth and reduce overshoot (reference tracking), e.g. to react faster to an impact with an obstacle or to simulate virtual environments properly. Also, the ideal controller would reject all disturbances and deliver the expected torque whenever the actuator is backdriven or not (torque accuracy). For example, this is particularly important for exoskeletons to ensure that the suit feels mechanically transparent to the user.

Three experiments are set to compare the performance of the controllers: a primary backdriving test to assess the effect of the dither, then a reference tracking test and finally a torque accuracy test. The experimental setup is shown in Fig.~\ref{fig_TestBench} and comprises an incremental magnetic encoder for the master piston displacement and speed (RLM2HDA12BB15B00, RLS, \SI{0.5}{\micro\metre} resolution), an optical encoder for the joint displacement (E5-5000-315-IE-S-D-G-1, US Digital) and pressure transducers (PX3AN1BH667PSAAX, Honeywell) for the master and slave cylinders. The joint torque is computed based on the slave pressure of the controlled line.

Next subsections describe the experimental validation of the controllers. Main results are summarized in table~\ref{table_ResultsSummary_}.

\begin{table*}[t]
\begin{threeparttable}
\vspace{10pt}
	\centering
	\caption{Results comparison}
	\label{table_ResultsSummary_}
		\begin{tabular}{ c c c c c c c }
			\hline
			Controller & Bandwidth & Rising time & Overshoot & 1 \si{Hz} torque deviation & 1 \si{Hz} torque deviation & 5 \si{Hz} torque deviation\\ 
			type\tnote{*} & & 63\% & & for 0 \si{N.m} command & for 10 \si{N.m} command & for 10 \si{N.m} command\tnote{**} \\
			 & (\si{Hz}) & (\si{ms}) & (\%) & (\si{N.m}) & (\si{N.m}) & (\si{N.m}) \\ \hline
			\hline
			\textbf{Open-loop (baseline)} & \textbf{25} & \textbf{16.6} & \textbf{34} & \textbf{0.60} & \textbf{2.4} & \textbf{4.2} \\ \hline 
			Open-loop + friction compens. & 25 & 15.8 & 38 & 0.38 & 1.2 & 5.0  \\ \hline 
			Master pressure PID & 11 & 17.2 & 10 & 0.17 & 0.5 & 3.1 \\ \hline 
			Slave pressure PID & 3 & 22.2 & 2 & 0.23 & 0.5 & 3.1 \\ \hline 
			State feedback LQGI & 34 & 14.4 & 19 & 0.24 & 0.6 & 1.8 \\ \hline 
		\end{tabular}
		\begin{tablenotes}
             \item[*] All controllers except for the baseline one use the dither signal.
             \item[**] These results are simulated here due to limitations to backdrive the arm consistently at high frequency (other results are experimental data).
        \end{tablenotes}
        \end{threeparttable}
\end{table*}


\subsection{Dither effect}
\begin{figure}%
    \centering
    \subfloat[\centering]{{\includegraphics[width=0.22\textwidth]{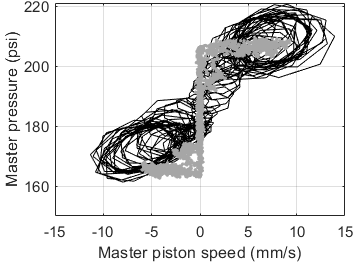}}}%
    \qquad
    \subfloat[\centering]{{\includegraphics[width=0.22\textwidth]{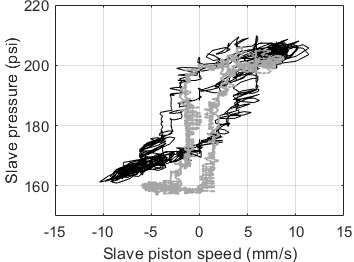}}}%
    \caption{Pressure variation due to backdriving at 1~\si{Hz} when dither is active (solid black) and when it is not (gray dot) for 1310~\si{kPa} (190~\si{psi}) nominal pressure~: a) master pressure versus master piston speed; b) slave pressure versus slave piston speed. Data is filtered at 200~\si{Hz}.}%
    \label{fig_DitherEffectTests}%
\end{figure}

Fig.~\ref{fig_DitherEffectTests}a shows master pressure variation when backdriving the end effector at 1~\si{Hz} for high nominal constant pressure. As expected, nonlinear friction added by ball screws generates sudden pressure variation around 0~\si{mm/s}. Consequently, low-speed position tracking with the robotic arm would lack stability \cite{chen_ballscrew_2000}. Some slave pressure hysteresis is seen in Fig.~\ref{fig_DitherEffectTests}b which depends on backdriving frequency and is due to the compliance between the disturbance source (at master cylinder side) and the slave cylinder. 

The dither signal changes the pressure behavior. In fact, one drawback is the induced 150~\si{Hz} cyclic master pressure variation. However, Fig.~\ref{fig_DitherEffectTests}b shows that vibrations are four times smaller at slave cylinder due to transmission damping. Also, when dither is active, the pressure variation is smoothed at low speed reversals in both figures. This is a desired effect since it will be more effective to reject these smooth low frequency disturbances with closed-loop feedback.

\subsection{Reference tracking}
Performance metrics chosen for reference tracking are joint torque bandwidth, 63\% rising time and overshoot when the end effector is blocked. These metrics are relevant for direct interactions with environments and users, e.g. for haptics, force control, impacts and so forth. Note that environmental impedance can limit force bandwidth at the end effector \cite{veronneau_lightweight_2019}.

\begin{figure}[h!]
\centering
\includegraphics[width=0.45\textwidth]{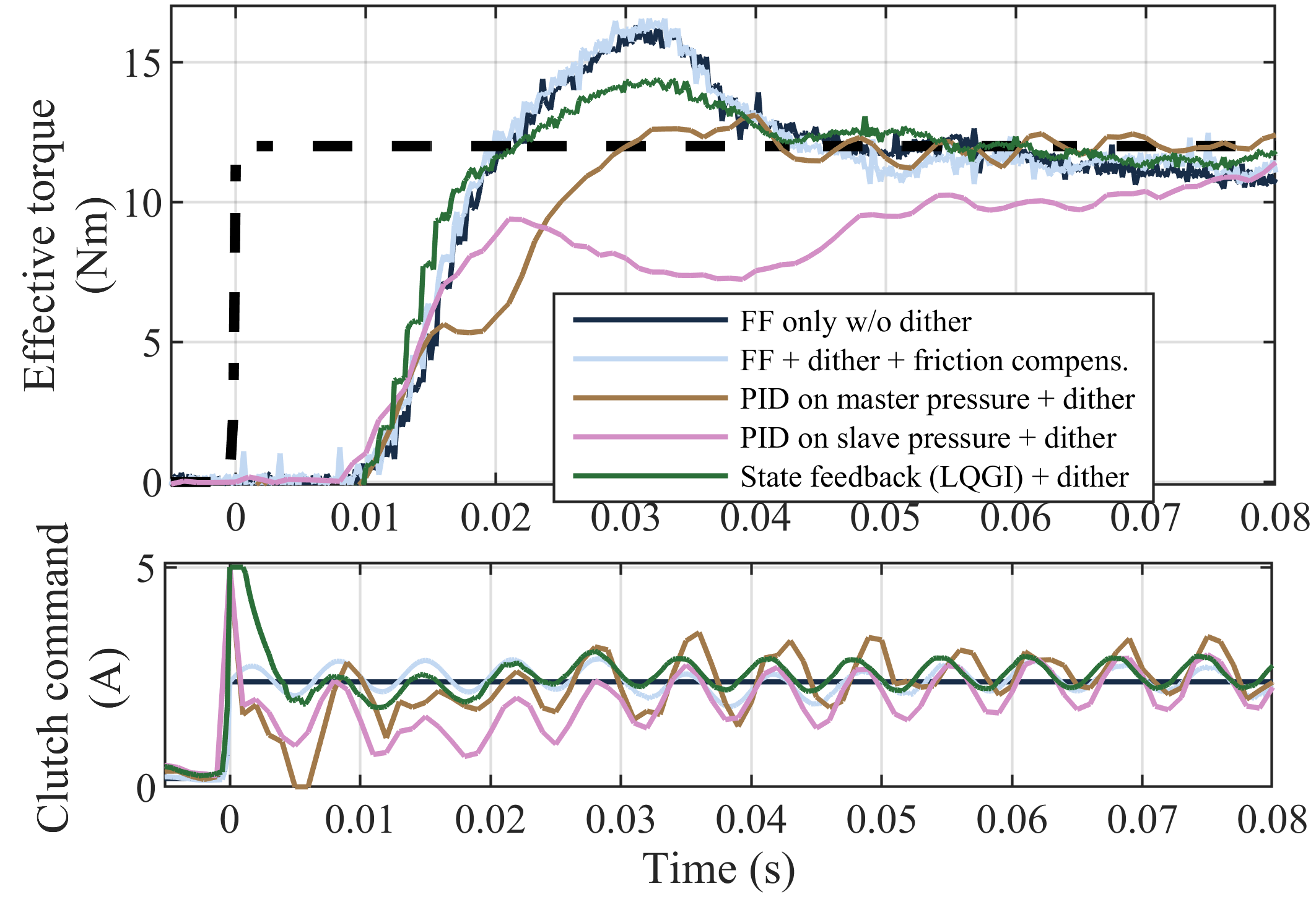}
\caption{Torque step response and clutch command for all controllers developed.}
\label{fig_StepResponsesControllers}
\end{figure}

\begin{figure}[h!]
\centering
\includegraphics[width=0.45\textwidth]{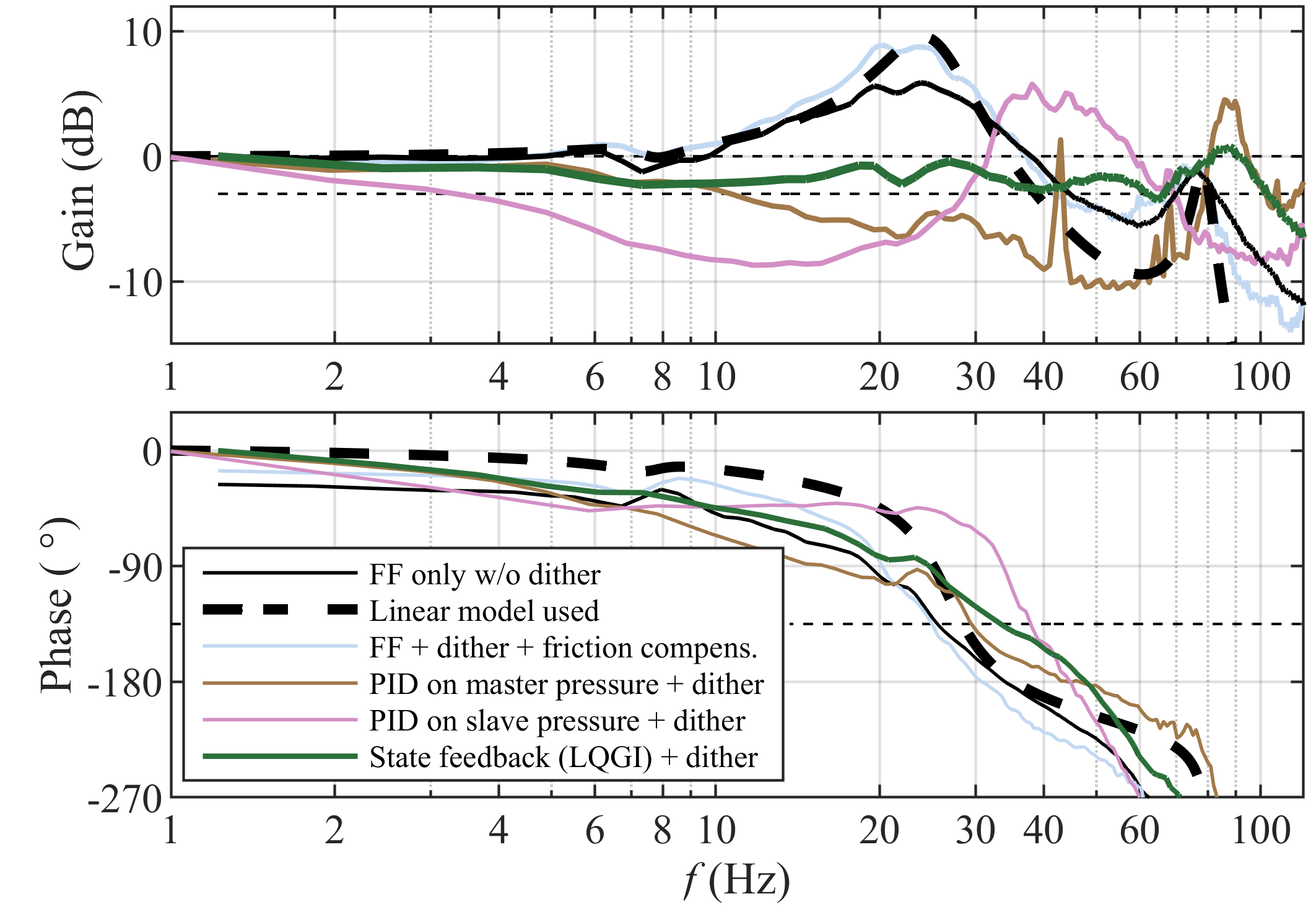}
\caption{Experimental torque frequency response ${{{P}_{\text{s}}}}/{{{P}_{\text{d}}}}$ with blocked end-effector for all controllers developed. Linear model is added for comparison.}
\label{fig_BodeResponsesControllers}
\end{figure}

For reference tracking, a 12~\si{N.m} step reference is sent followed by a 10~$\pm$~2~\si{N.m} chirp command. The torque step response for each controller is presented in Fig.~\ref{fig_StepResponsesControllers} and the frequency responses are presented in Fig.~\ref{fig_BodeResponsesControllers}. The torque bandwidth found is the minimum frequency at which the amplitude is -3~\si{dB} or the phase is -135$\degree$.

The baseline for comparing the control approaches is the open-loop controller (without dither). The performance metrics found in this case are a 25~\si{Hz} bandwidth, a 16.6~\si{ms} rising time and a 34\% overshoot due to transmission compliance. For the second controller, the dither and friction compensation have little effects on reference tracking. Nonetheless, the DC phase lag (at 1~\si{Hz}) is improved from -26$\degree$ to -16$\degree$. For the pressure feedback controllers, the overshoot is greatly reduced as desired. However, the 63\% rising time increases to 17.2~\si{ms} and to 22.2~\si{ms} for master and slave pressure feedback, respectively. Pressure feedback gains limit reference tracking speed, especially for slave pressure feedback. As discussed in section~\ref{subsection_PressureFeedbackCTRL}, higher-order controllers could be developed to increase bandwidth near the open-loop controller.

Finally, the proposed LQGI controller increases reference tracking performances with a 34~\si{Hz} torque bandwidth, a 14.4~\si{ms} rising time and with less overshoot compared to the baseline. Also, the frequency response is flatter, i.e. that torque is $\pm$~30\% accurate up to 34~\si{Hz}. Above 34~\si{Hz}, the phase shift exceeds -135$\degree$.

 Hence, the model-based approach reaches high-bandwidth torque fidelity by attenuating the actuator command around the natural frequencies of the hydrostatic transmission while reducing the phase lag due to transmission compliance and MR clutch dynamics.

\subsection{Torque accuracy}
To assess torque accuracy, the performance metric is the peak deviation torque when the elbow joint is backdriven for five cycles at 1~\si{Hz} with a 0~\si{N.m} nominal command and with a 10~\si{N.m} command. The DC pressure of each line is 205~\si{kPa} (30~\si{psi}). The effective torque is compared with the reference. At this frequency, the reference tracking error is mainly due to ball screw friction, not to the actuator and transmission damping and inertia.

For the 0~\si{N.m} nominal command tests, the baseline deviation is 0.6~\si{N.m}. The friction compensation controller with dither improves the metric by 37\% while the pressure feedback controllers result in a 72\% improvement. The torque deviation reduction for the LQGI controller is like the PID controllers with 60\% enhancement. 

For a 10~\si{N.m} nominal command, the deviation is 2.4~\si{N.m} for the baseline open-loop test. The improvement for each controller is similar to the 0~\si{N.m} command tests. Lower torque deviation could be possible with more aggressive integral gains for the LQGI controller but it penalized reference tracking performances.

For a deeper comparison, simulations were conducted at 5~\si{Hz} backdriving frequency. At this frequency, transmission damping and inertia have more effects on torque deviation. The linear model presented in section~\ref{section_modeling} was used along with MR pure delay and ball screw stick-slip Coulomb friction. Simulated results show that friction compensation is no longer working. Indeed, transmission hysteresis (Fig.~\ref{fig_DitherEffectTests}b) becomes so high at 5~\si{Hz} that the simplified friction model introduces  out of phase commands. The LQGI controller improves much better torque deviation than pressure feedback. In fact, at high backdriving frequency, state feedback can reduce the effect of the transmission's damping and inertia.

Overall, most controllers in this paper showed limitations to improve backdriving torque accuracy while having good reference tracking performances. In contrast, the LQGI controller improved torque fidelity in both aspects.
\section{Conclusion}
Robots working in uncontrolled environments and interacting with humans should deliver the expected forces fast even in presence of unknown disturbances. In this paper, four control strategies were developed and compared to improve torque fidelity of an MR-hydrostatic actuator.

The experiments highlighted limitations of open-loop controllers and pressure-based PID controllers. Performance of the latters was limited by stability issues. Still, a dither strategy at a specific frequency effectively smoothed the discontinuous effect of ball screw friction which improved disturbance rejection of closed-loop controllers. Also, a simple friction compensation technique reduced torque deviation but requires precise position sensors and is not working at high backdriving frequency. All performance metrics were improved with an LQGI approach based on a linear model and a few sensor inputs. Overall, this approach boosts the torque fidelity by increasing the bandwidth from 25~\si{Hz} to 34~\si{Hz}, by reducing overshoot from 34\% to 19\% and by reducing the torque deviation when backdriven from 24\% to 6\% of nominal command. 

Other hydrostatic architectures with different actuator technologies could benefit from the state feedback strategy. More assessments and improvements are worth considering for model-based control of hydrostatic transmissions in robotics. Future work includes gain scheduling techniques to compensate for the stiffening of hydrostatic transmissions when pressure increases. The performance, stability and robustness of the free output configuration when the device is carried by a user should also be assessed.

\bibliographystyle{IEEEtran}
\bibliography{Zotero}

\end{document}